\pdfoutput=1
\PassOptionsToPackage{margin=0in}{geometry}
\documentclass[11pt]{article}

\usepackage[preprint]{acl}

\usepackage{times}
\usepackage{latexsym}

\usepackage[T1]{fontenc}

\usepackage[utf8]{inputenc}

\usepackage{microtype}

\usepackage{inconsolata}

\usepackage{graphicx}

\usepackage{amsmath}
\usepackage{amsfonts}
\usepackage{subcaption}

\usepackage{listings}
\usepackage[frozencache,cachedir=.]{minted}
\usepackage{tcolorbox}
\usepackage{caption}
\usepackage{tabularx}

\newcommand{\subtwosection}[1]{\noindent\textbf{#1}}

\title{Enhance the Robustness in Text-Centric Multimodal Alignments}

\author{Ting-Yu Yen \\ National Taiwan University \\        
        \texttt{r11922042@ntu.edu.tw} \And
        Yun-Da Tsai \\ National Taiwan University \\ \texttt{f08946007@csie.ntu.edu.tw} \And 
        Keng-Te Liao \\ National Taiwan University \\ \texttt{d05922001@ntu.edu.tw} \AND
        Shou-De Lin \\ National Taiwan University \\ \texttt{sdlin@csie.ntu.edu.tw}}

\begin{document}
\maketitle

\begin{abstract}
Converting different modalities into general text, serving as input prompts for large language models (LLMs), is a common method to align multimodal models when there is limited pairwise data. This text-centric approach leverages the unique properties of text as a modality space, transforming diverse inputs into a unified textual representation. This enables downstream models to effectively interpret various modal inputs. This study assesses the quality and robustness of multimodal representations in the presence of missing entries, noise, or absent modalities, revealing that current text-centric alignment methods compromise downstream robustness. To address this issue, we propose a new text-centric approach that achieves superior robustness compared to previous methods across various modalities in different settings. Our findings highlight the potential of this approach to enhance the robustness and adaptability of multimodal representations, offering a promising solution for dynamic and real-world applications.
\end{abstract}

\section{Introduction}
\label{sec:intro}

Text-centric multimodal alignment methods have emerged as a powerful approach for integrating multimodal information by converting diverse data types into text. This technique leverages text as a universal modality, enabling large language models (LLMs) to process and understand visual, auditory, and other forms of data. By transforming non-textual information into textual descriptions, these methods enhance the capability of LLMs to generate contextually rich responses. For example, LLaVA~\cite{liu2023llava} uses expert models to generate captions and descriptions from images, which are then input to GPT-4 to create vision-text instruction-following data, circumventing the difficulty of collecting such data directly.

\begin{figure}[t]
\centering
  \centering
  \includegraphics[width=0.95\linewidth]{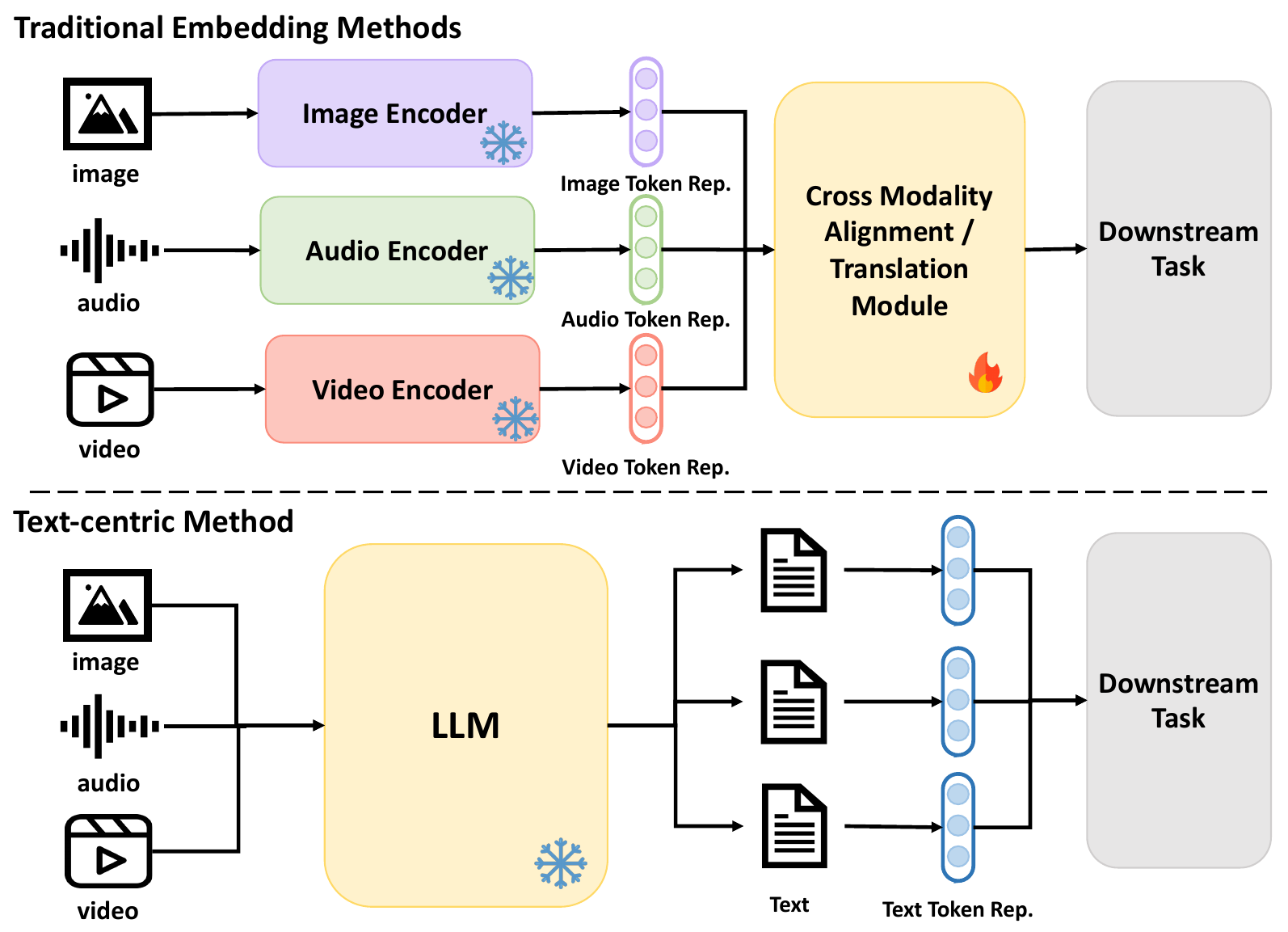}
  \caption{Text-centric multimodal alignment, which converts different modalities into text to serve as input prompts for LLMs, is a common method for aligning large multimodal language models when pairwise multimodal data is limited.}
  \label{fig:text-centric}
\end{figure}

\begin{figure*}[t!]
\centering
  \centering
  \includegraphics[width=0.95\linewidth]{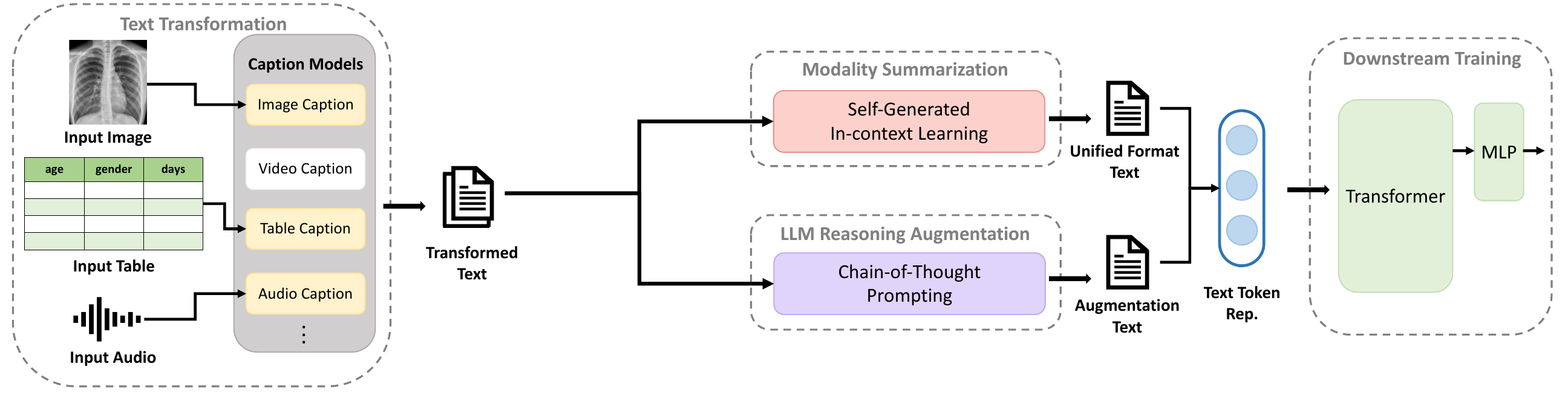}
  \caption{Each raw input modality is transformed into text representations using a corresponding foundation model. Following modality summarization and LLM reasoning are applied in parallel. Finally, the output texts are concatenated as the input to a transformer model for downstream prediction. The inference phase follows a similar pattern. We apply a one-shot in-context learning approach to adapt the linguistic style as anticipated during training.}
  \label{fig:method}
\end{figure*}

\nocite{tsai2023differential}

Recent studies~\cite{wang2023too} discovered that Vision LLMs trained on synthetically generated high-quality captions can suffer from model collapse~\cite{robinson2021can} (focus solely on a subset of
the modalities). This occurs due to captioning collapse~\cite{vinyals2015show,wang2020overview} and the one-to-many problem~\cite{young2014image}, where similar captions for different images limit diversity and potentially harm downstream model training. This leads to less robust multimodal representations and issues like modality collapse in generative models.

\nocite{tsai2024toward}

Real-world multimodal data often contains missing entries, noise, or absent modalities, necessitating robust models for accurate predictions. While text-centric methods are widely adopted, our work is the first to explore robustness and collapse issues in text-centric modality alignment. We evaluate robustness using an approach similar to the MULTIBENCH framework~\cite{liang2021multibench}, simulating real-world conditions with varying levels of noise and imperfections in the input modalities. Our results show that transforming various modalities into text can lead to reduced robustness of multimodal representations, as evidenced by lower accuracy in downstream tasks.

To address this robustness issue, we propose an approach to further enhance text-centric multimodal alignment. After converting different input modalities into text using expert foundation models, we align these modalities within a similar semantic space and enhance interactions by applying summarization across modalities, combining textual representations into a unified form. This is followed by modality reasoning, utilizing large-scale external knowledge for data augmentation.

Through experiments, we demonstrate that our enhancement can significantly improve the modality robustness. Qualitative analysis also shows that modality summarization and reasoning augmentation with LLMs offer significant advantages: 1) recovering dropped or corrupted information, 2) transforming implicit relationships into explicit text descriptions, and 3) compensating missing information using LLMs as external knowledge sources.

\nocite{tsai2024handling}

Our contributions are summarized as follows:
\begin{itemize}
    \item We are the first to investigate modality robustness in text-centric alignment methods, revealing their inherent lack of robustness.
    \item We propose an enhancement for text-centric alignment that demonstrates effective performance, outperforming the best baseline by 15.2\% in experiments on real-world datasets.
    \item We provide a qualitative analysis illustrating how large language models (LLMs) strengthen the robustness of textual representations in multimodal alignment.
\end{itemize}

\section{Related Work}
\label{sec:related}

\subsection{Text-centric Multimodal Alignment}
Recent studies highlight the effectiveness of text-centric alignment. LLaVA~\cite{liu2023llava} uses GPT-4 to generate image captions, while VideoChat-Text~\cite{li2023videochat} encodes video content into text. In the medical field, OphGLM~\cite{gao2023ophglm} and ChatCAD~\cite{wang2023chatcad} convert medical images into diagnostic reports. TAMML~\cite{tsai2024text} transforms various input modalities into text, improving performance on diverse modalities. These methods depend on the quality of the transformed text but offer a straightforward approach to multimodal integration.

\subsection{Challenges in Multimodal Learning}
Multimodal training faces several challenges. Modality Robustness~\cite{ma2022multimodal} deals with noise and missing data in modalities. Modality Competition~\cite{huang2022modality} involves different modalities competing, leading to better performance by the best uni-modal network. {Modality Collapse~\cite{javaloy2022mitigating} occurs when generative models neglect some modalities. Modality Forgetting~\cite{peng2023sparse} happens when conflicting optimization directions cause knowledge loss.
Text-centric alignment methods face similar issues. Vision LLMs trained on synthetic high-quality captions often experience model collapse~\cite{wang2023too}, explained by captioning collapse~\cite{vinyals2015show,wang2020overview} and the one-to-many problem~\cite{young2014image}, where fixed captions limit output diversity and lead to trivial solutions.

\begin{figure*}[t]
\vspace{-1.4cm}
  \centering
  \includegraphics[width=0.35\linewidth]{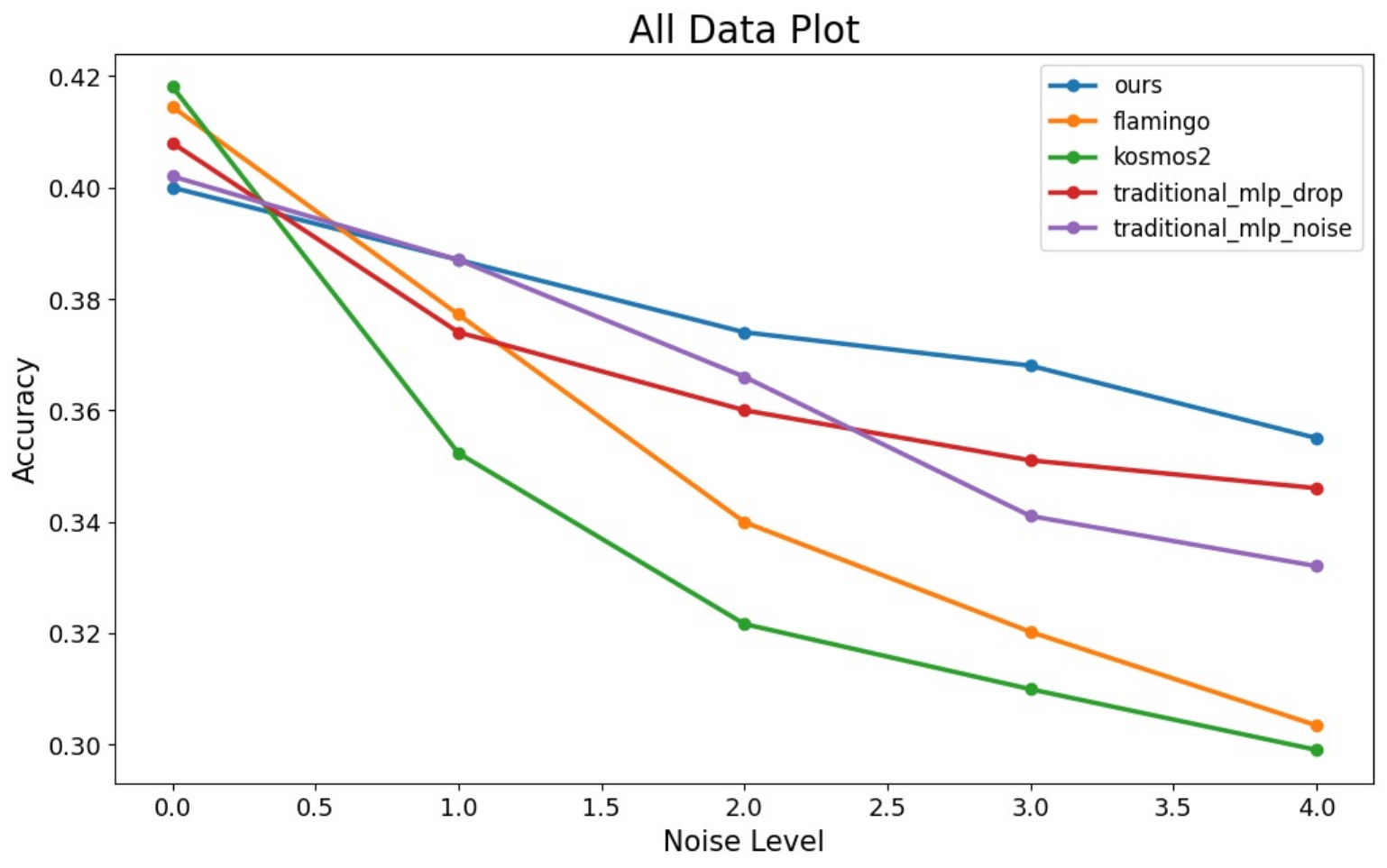}
  \includegraphics[width=0.35\linewidth]{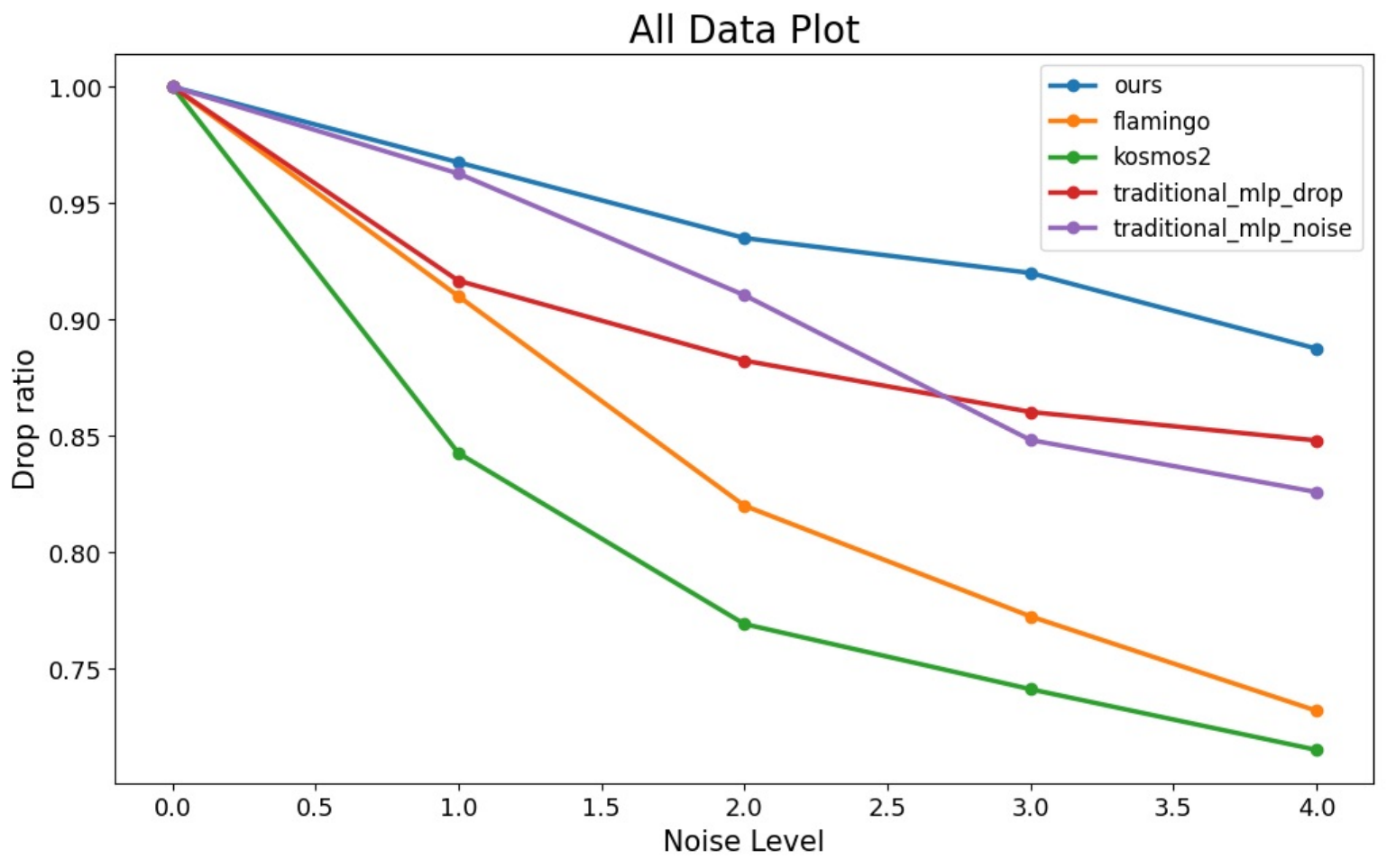}
  \caption{To evaluate robustness with noise applied to all three modalities, we analyzed accuracy (left) and drop ratio (right). Both metrics indicate that the text-centric method demonstrates stronger robustness and better resistance to noise compared to other baselines as the noise level increases.}
  \label{fig:acc_drop_all}
\end{figure*}

\begin{figure*}[t]
  \centering
  \includegraphics[width=1.0\linewidth]{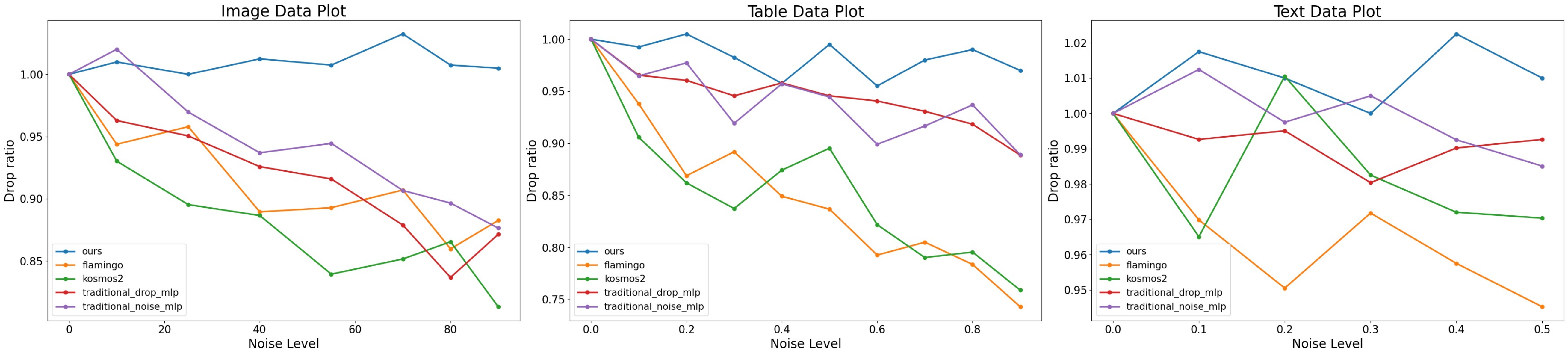}
  \caption{Drop ratio when noise is applied to modalities separately - Image (left) and Table (center) and Text (right). Text-centric methods outperforms all baselines and remain nearly no declined as the noise level increase.}
  \label{fig:acc_drop_image}
\end{figure*}



\section{Methodology}
\label{sec:method}

\subsection{Text Transformation}
We use text-based representations for various modalities to enhance model robustness. Inputs are converted into standardized text formats, reducing modality-specific adaptations and data complexity, which captures essential information while filtering out noise. For image data, we use a SOTA image captioning model to generate detailed textual descriptions. Textual data remains unchanged to preserve its integrity. For tabular data, we apply a serialization method from TabLLM~\cite{hegselmann2023tabllm}, structured as "The column name is values," which outperforms zero-shot learning in LLMs. The transformed texts from each modality are then merged for further processing.

\subsection{Modality Summarization}
To address syntactic and semantic gaps between transformed texts, we extend similar linguistic styles to all modalities, improving information quality and removing redundancies. Using LLMs, we summarize the modalities in two phases: first, we collect samples using prompts that guide LLMs to merge information into concise summaries; second, we integrate these outputs with original prompts for in-context learning applied to subsequent samples.

\subsection{LLM Reasoning}
We utilize LLMs for reasoning with the Chain-of-Thought method~\cite{wei2022chain} and as large-scale external knowledge sources for data augmentation~\cite{KDA_2023}. By assigning prediction tasks with clear instructions and examples, LLMs analyze and augment textual inputs, generating predictions and detailed explanations to enhance the data.

\nocite{tsai2023rtlfixer}

\section{Experiment}
\label{sec:exp}

\subsection{Experiment Setup}

\subtwosection{Dataset}
\label{sec:exp:dataset}
In our experiments, we adopted the \textit{PetFinder}~\cite{petfinder-adoption-prediction} classification dataset, which is a composite of 3 modalities: Text, Image, and Tabular.

\nocite{da2022fast}

\subtwosection{Baselines}
We selected two SOTA MLLMs for robustness comparison: Kosmos-2~\cite{peng2023kosmos} and Flamingo~\cite{alayrac2022flamingo}. After fine-tuning, these were followed by using an MLP as the backbone model to generate predictions. Additionally, we employed robust representation learning techniques in the MLP, including noise injection and dropout as our baselines.

\subsection{Evaluation}
\label{sec:exp:eval}
\nocite{tsai2021toward}

\subtwosection{Evaluation Protocol}
To evaluate the robustness of our models, we adopted the methodologies outlined in MULTIBENCH~\cite{liang2021multibench}. The first is modality-specific imperfections applied to each modality.	The second is multimodal imperfections, where noise is applied to all modalities. For images, we introduced Gaussian noise at five different levels from 10\% to 90\%. For text descriptions, we randomly dropped words with five levels of probability from 10\% to 50\%. For table data, we randomly dropped column features with probabilities from 10\% to 90\%.

\subtwosection{Evaluation Metric}
Following our evaluation protocols designed to mimic the modality-specific and multimodal imperfections described in MULTIBENCH, we evaluate both \textbf{Accuracy} under imperfections (relative robustness) and the \textbf{Drop ratio} of accuracy when imperfections are introduced (effective robustness) of our models.


\subsection{Results}
Figure~\ref{fig:acc_drop_all} shows that our method has the lowest drop ratio under all-modality conditions, with outperforming other baselines with the highest noise levels. Our method experienced only an 11.3\% drop, significantly better than the best-performing baseline at 15.2\% and far better than the text-centric alignment using Kosmos-2, which dropped 28.5\%. This indicates that even MLLMs pre-trained with extensive datasets cannot withstand the impact of noise.
Figure~\ref{fig:acc_drop_image} reveals that our approach maintains a significant advantage while modality-specific noise is applied across all modalities, the highest drop ratio of our approach being only 4.5\%, much lower than the best baseline's 11.2\%.

\begin{figure}[t]
  \centering
  \includegraphics[width=0.9\linewidth]{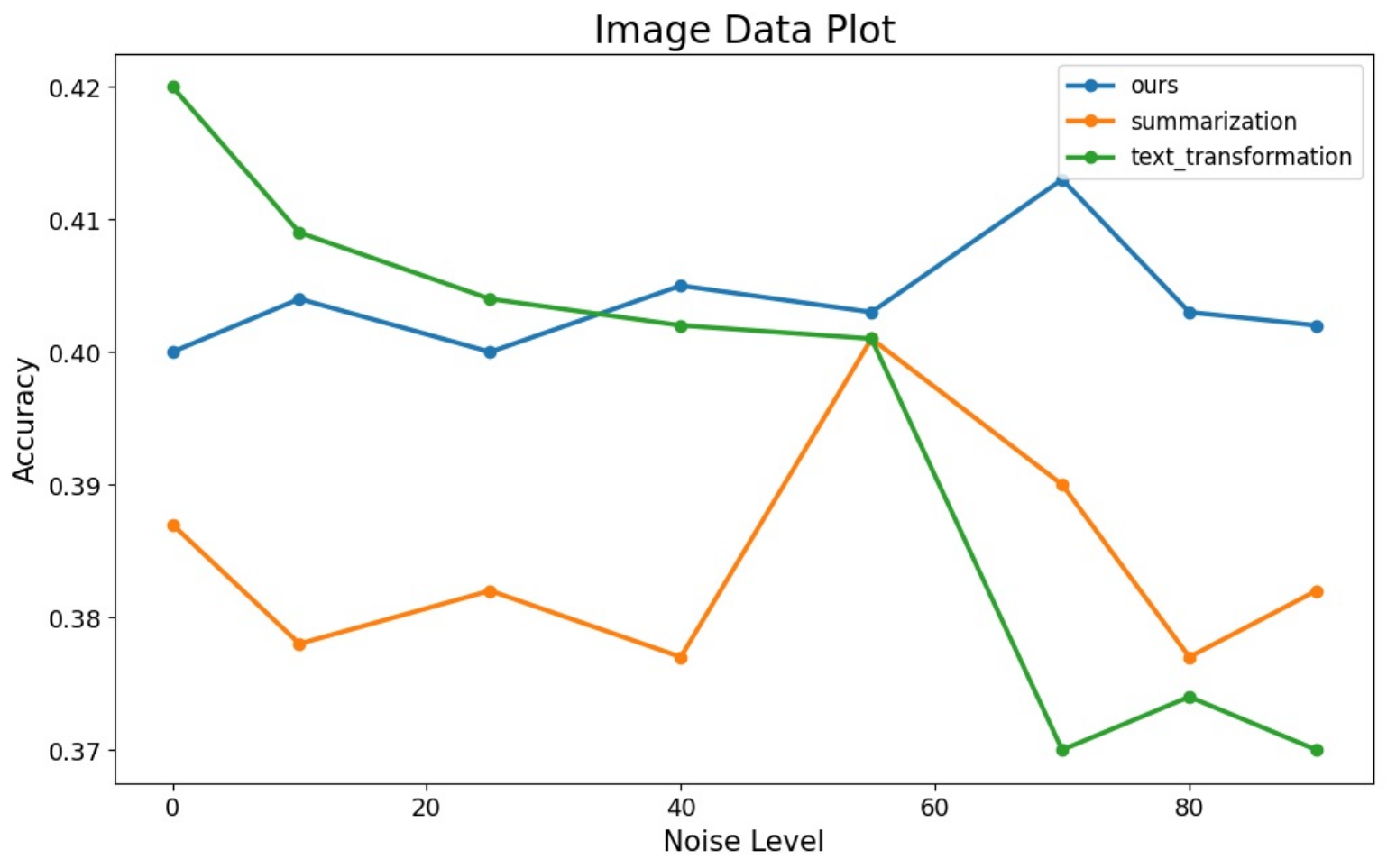}
  \caption{Component ablation studies suggest that removing modality summarization and reasoning greatly reduces the performance of the our method under increased noise levels. This indicates the critical effectiveness of two major components in our method.}
  \label{fig:component_ablation}
\end{figure}

\section{Ablation Study}
\label{sec:ablation}

\subtwosection{Method Component Ablation}
In this study, we focus on the two major components of our method: Modality Summarization and Reasoning Augmentation.
The results in Figure~\ref{fig:component_ablation} suggest that without modality summarization and reasoning augmentation, the performance drops significantly with the drop rate decreasing from 11.9\% to 1.9\% at maximum noise levels. Additionally, our approach improves accuracy by approximately 8\% over text transformation and outperforms summarization alone by approximately 5\%. This indicates the critical effectiveness of these two major components in our method.

\subtwosection{LLM Ablation}
We adopted different LLMs to test whether robustness is consistent across various LLMs and how well it is sustained with different model types and sizes. Table~\ref{tab:ablation:llm} shows that GPT-4o offers the best performance among all LLMs. However, the impact of model type and size is minor, with a maximum difference of around 2\% in accuracy. We conclude that the robustness of our method is transferable between models.



\begin{table}[t]
\centering
\begin{tabular}{lccc}
\hline
Model & Image & Text & Table \\
\hline
GPT-4O & \textbf{0.4038} & \textbf{0.3990} & \textbf{0.3931} \\
GPT-3.5-turbo & 0.3975 & 0.3936 & 0.3901 \\
Mixtral8x7b & 0.3938 & 0.3981 & 0.3882 \\
\hline
\end{tabular}
\caption{Ablation study on the impact of different LLMs. GPT-4o offers the best performance, but the impact between LLMs is not substantial and, at max, $\sim$2\% accuracy.}
\label{tab:ablation:llm}
\end{table}

\section{Qualitative Analysis and Findings}
\label{sec:analysis}

\begin{figure}[ht!]
\begin{tcolorbox}[width=1.0\linewidth, halign=left, colframe=black, colback=white, boxsep=0.01mm, arc=1.5mm, left=2mm, right=2mm, boxrule=0.5pt]\footnotesize

\textbf{Text with noise}\\ 
\textcolor{gray}{The type of pet is Dog. The name of pet is Filo. \textcolor{red}{The age of pet when listed (in months) is 78.}The color 1 of pet is Brown. The fur length of pet is Short.  The primary breed of pet is Mixed Breed. \textcolor{red}{The secondary breed of pet (if pet is of mixed breed) is Unknown.}
}

\textbf{Summarization}\\ 
\textcolor{blue}{This pet is a gentle and handsome 6-year-old mixed breed dog named Filo. He has a light brown, short coat and soft brown eyes.} 

\end{tcolorbox}
\caption{The tabular data has dropped the color and fur length column (gray). However, it was recovered (blue) after applying modality summarization with LLM that compensate the information from input image.}
\label{fig:analysis-table}
\end{figure}

\begin{figure}[ht!]
\begin{tcolorbox}[width=1.0\linewidth, halign=left, colframe=black, colback=white, boxsep=0.01mm, arc=1.5mm, left=2mm, right=2mm, boxrule=0.5pt]\footnotesize

\textbf{Text with noise}\\ 
\textcolor{red}{Jack experienced it Jack dog even is He recovering but special care skin which hair to Vet this was to Hopefully lover who take good dog a}

\textbf{Summarization}\\ 
\textcolor{blue}{Although the profile write-up for Jack Jack is somewhat fragmented, it indicates that he has experienced some challenges but is on the road to recovery} 

\textbf{Reasoning}\\ 
\textcolor{cyan}{3. **Health Condition**: Jack Jack has a minor injury and is recovering. Potential adopters might be hesitant to take on a pet that requires special care, even if the injury is minor.}

\end{tcolorbox}
\caption{This example demonstrates that summarization and reasoning can compensate for noisy text input, transforming it into clear descriptions.}
\label{fig:analysis-text}
\end{figure}

\subtwosection{LLMs recover lost data from other modalities}
Figure~\ref{fig:analysis-table} shows that even when my input loses significant information, we can still obtain information from other modalities (see Appendix~\ref{sec:appendix} for detailed input) to recover the missing data.

\subtwosection{LLMs compensate missing information with knowledge and transform implicit relations into explicit text description by reasoning.}
When the input words are dropped making sentences incoherent, and when there is no relevant information in other modality, Figure~\ref{fig:analysis-text} demonstrates how our method uses LLM’s knowledge to recover the original words through summarization and even learns the potential meaning behind the information through reasoning.

\section{Conclusion}
\label{sec:conclusion}
This study evaluates the robustness of text-centric multimodal alignment and reveals its less robustness compared to others. We propose a new text-centric alignment approach that outperforms the baselines and exhibits strong resistance to noise. Ablation studies indicate that modality summarization and reasoning augmentation are critical components for enhancing robustness. Additionally, the robustness of our method is highly transferable across different LLMs. These findings provide valuable insights for developing more robust multimodal alignment approaches.

\section{Limitation}
\label{sec:limitation}
One of the key limitations of our study is the inherent randomness of the LLM text generation. Due to cost constraints, we only performed one run for each of our experiments. While this approach provides a general indication of performance, it may not fully capture the variability and could lead to less accurate conclusions. More extensive experimentation with a larger number of runs would be necessary to achieve a higher degree of confidence in the results. In addition, we cannot guarantee to reproduce the results on the closed-source LLMs.

\section*{Use of AI Assistants}
ChatGPT was utilized to refine paper writing. The authors paid careful attention to ensuring that AI-generated content is accurate and aligned with the author's intentions.

\bibliographystyle{unsrtnat}
\bibliography{custom}

\begin{thebibliography}{27}
\providecommand{\natexlab}[1]{#1}
\providecommand{\url}[1]{\texttt{#1}}
\expandafter\ifx\csname urlstyle\endcsname\relax
  \providecommand{\doi}[1]{doi: #1}\else
  \providecommand{\doi}{doi: \begingroup \urlstyle{rm}\Url}\fi

\bibitem[Liu et~al.(2023)Liu, Li, Wu, and Lee]{liu2023llava}
Haotian Liu, Chunyuan Li, Qingyang Wu, and Yong~Jae Lee.
\newblock Visual instruction tuning, 2023.

\bibitem[Tsai et~al.(2023{\natexlab{a}})Tsai, Tsai, and Lin]{tsai2023differential}
Yun-Da Tsai, Tzu-Hsien Tsai, and Shou-De Lin.
\newblock Differential good arm identification.
\newblock \emph{arXiv preprint arXiv:2303.07154}, 2023{\natexlab{a}}.

\bibitem[Wang et~al.(2023{\natexlab{a}})Wang, Lin, Zhang, Lei, and Shou]{wang2023too}
Alex~Jinpeng Wang, Kevin~Qinghong Lin, David~Junhao Zhang, Stan~Weixian Lei, and Mike~Zheng Shou.
\newblock Too large; data reduction for vision-language pre-training.
\newblock In \emph{Proceedings of the IEEE/CVF International Conference on Computer Vision}, pages 3147--3157, 2023{\natexlab{a}}.

\bibitem[Robinson et~al.(2021)Robinson, Sun, Yu, Batmanghelich, Jegelka, and Sra]{robinson2021can}
Joshua Robinson, Li~Sun, Ke~Yu, Kayhan Batmanghelich, Stefanie Jegelka, and Suvrit Sra.
\newblock Can contrastive learning avoid shortcut solutions?
\newblock \emph{Advances in neural information processing systems}, 34:\penalty0 4974--4986, 2021.

\bibitem[Vinyals et~al.(2015)Vinyals, Toshev, Bengio, and Erhan]{vinyals2015show}
Oriol Vinyals, Alexander Toshev, Samy Bengio, and Dumitru Erhan.
\newblock Show and tell: A neural image caption generator.
\newblock In \emph{Proceedings of the IEEE conference on computer vision and pattern recognition}, pages 3156--3164, 2015.

\bibitem[Wang et~al.(2020)Wang, Zhang, and Yu]{wang2020overview}
Haoran Wang, Yue Zhang, and Xiaosheng Yu.
\newblock An overview of image caption generation methods.
\newblock \emph{Computational intelligence and neuroscience}, 2020\penalty0 (1):\penalty0 3062706, 2020.

\bibitem[Young et~al.(2014)Young, Lai, Hodosh, and Hockenmaier]{young2014image}
Peter Young, Alice Lai, Micah Hodosh, and Julia Hockenmaier.
\newblock From image descriptions to visual denotations: New similarity metrics for semantic inference over event descriptions.
\newblock \emph{Transactions of the Association for Computational Linguistics}, 2:\penalty0 67--78, 2014.

\bibitem[Tsai et~al.(2024{\natexlab{a}})Tsai, Liow, Siang, and Lin]{tsai2024toward}
Yun-Da Tsai, Cayon Liow, Yin~Sheng Siang, and Shou-De Lin.
\newblock Toward more generalized malicious url detection models.
\newblock In \emph{Proceedings of the AAAI Conference on Artificial Intelligence}, volume~38, pages 21628--21636, 2024{\natexlab{a}}.

\bibitem[Liang et~al.(2021)Liang, Lyu, Fan, Wu, Cheng, Wu, Chen, Wu, Lee, Zhu, et~al.]{liang2021multibench}
Paul~Pu Liang, Yiwei Lyu, Xiang Fan, Zetian Wu, Yun Cheng, Jason Wu, Leslie Chen, Peter Wu, Michelle~A Lee, Yuke Zhu, et~al.
\newblock Multibench: Multiscale benchmarks for multimodal representation learning.
\newblock \emph{arXiv preprint arXiv:2107.07502}, 2021.

\bibitem[Tsai and Lin(2024)]{tsai2024handling}
Yun-Da Tsai and Shou-De Lin.
\newblock Handling concept drift in non-stationary bandit through predicting future rewards.
\newblock In \emph{Pacific-Asia Conference on Knowledge Discovery and Data Mining}, pages 161--173. Springer, 2024.

\bibitem[Li et~al.(2023)Li, He, Wang, Li, Wang, Luo, Wang, Wang, and Qiao]{li2023videochat}
KunChang Li, Yinan He, Yi~Wang, Yizhuo Li, Wenhai Wang, Ping Luo, Yali Wang, Limin Wang, and Yu~Qiao.
\newblock Videochat: Chat-centric video understanding.
\newblock \emph{arXiv preprint arXiv:2305.06355}, 2023.

\bibitem[Gao et~al.(2023)Gao, Deng, Niu, Rong, Chen, Gong, Zhang, Xiao, Li, Cao, et~al.]{gao2023ophglm}
Weihao Gao, Zhuo Deng, Zhiyuan Niu, Fuju Rong, Chucheng Chen, Zheng Gong, Wenze Zhang, Daimin Xiao, Fang Li, Zhenjie Cao, et~al.
\newblock Ophglm: Training an ophthalmology large language-and-vision assistant based on instructions and dialogue.
\newblock \emph{arXiv preprint arXiv:2306.12174}, 2023.

\bibitem[Wang et~al.(2023{\natexlab{b}})Wang, Zhao, Ouyang, Wang, and Shen]{wang2023chatcad}
Sheng Wang, Zihao Zhao, Xi~Ouyang, Qian Wang, and Dinggang Shen.
\newblock Chatcad: Interactive computer-aided diagnosis on medical image using large language models.
\newblock \emph{arXiv preprint arXiv:2302.07257}, 2023{\natexlab{b}}.

\bibitem[Tsai et~al.(2024{\natexlab{b}})Tsai, Yen, Guo, Li, and Lin]{tsai2024text}
Yun-Da Tsai, Ting-Yu Yen, Pei-Fu Guo, Zhe-Yan Li, and Shou-De Lin.
\newblock Text-centric alignment for multi-modality learning.
\newblock \emph{arXiv preprint arXiv:2402.08086}, 2024{\natexlab{b}}.

\bibitem[Ma et~al.(2022)Ma, Ren, Zhao, Testuggine, and Peng]{ma2022multimodal}
Mengmeng Ma, Jian Ren, Long Zhao, Davide Testuggine, and Xi~Peng.
\newblock Are multimodal transformers robust to missing modality?
\newblock In \emph{Proceedings of the IEEE/CVF Conference on Computer Vision and Pattern Recognition}, pages 18177--18186, 2022.

\bibitem[Huang et~al.(2022)Huang, Lin, Zhou, Yang, and Huang]{huang2022modality}
Yu~Huang, Junyang Lin, Chang Zhou, Hongxia Yang, and Longbo Huang.
\newblock Modality competition: What makes joint training of multi-modal network fail in deep learning?(provably).
\newblock In \emph{International Conference on Machine Learning}, pages 9226--9259. PMLR, 2022.

\bibitem[Javaloy et~al.(2022)Javaloy, Meghdadi, and Valera]{javaloy2022mitigating}
Adri{\'a}n Javaloy, Maryam Meghdadi, and Isabel Valera.
\newblock Mitigating modality collapse in multimodal vaes via impartial optimization.
\newblock In \emph{International Conference on Machine Learning}, pages 9938--9964. PMLR, 2022.

\bibitem[Peng et~al.(2023{\natexlab{a}})Peng, Zhou, Zhou, Hartvigsen, Zhang, Wang, and Chen]{peng2023sparse}
Jie Peng, Kaixiong Zhou, Ruida Zhou, Thomas Hartvigsen, Yanyong Zhang, Zhangyang Wang, and Tianlong Chen.
\newblock Sparse moe as a new treatment: Addressing forgetting, fitting, learning issues in multi-modal multi-task learning.
\newblock 2023{\natexlab{a}}.

\bibitem[Hegselmann et~al.(2023)Hegselmann, Buendia, Lang, Agrawal, Jiang, and Sontag]{hegselmann2023tabllm}
Stefan Hegselmann, Alejandro Buendia, Hunter Lang, Monica Agrawal, Xiaoyi Jiang, and David Sontag.
\newblock Tabllm: Few-shot classification of tabular data with large language models.
\newblock In \emph{International Conference on Artificial Intelligence and Statistics}, pages 5549--5581. PMLR, 2023.

\bibitem[Wei et~al.(2022)Wei, Wang, Schuurmans, Bosma, Xia, Chi, Le, Zhou, et~al.]{wei2022chain}
Jason Wei, Xuezhi Wang, Dale Schuurmans, Maarten Bosma, Fei Xia, Ed~Chi, Quoc~V Le, Denny Zhou, et~al.
\newblock Chain-of-thought prompting elicits reasoning in large language models.
\newblock \emph{Advances in Neural Information Processing Systems}, 35:\penalty0 24824--24837, 2022.

\bibitem[Chen et~al.(2023)Chen, Li, Hong, Xu, Gu, Lan, Zhu, and Wang]{KDA_2023}
Haoxing Chen, Yaohui Li, Yan Hong, Zhuoer Xu, Zhangxuan Gu, Jun Lan, Huijia Zhu, and Weiqiang Wang.
\newblock Boosting audio-visual zero-shot learning with large language models.
\newblock \emph{arXiv preprint arXiv: 2311.12268}, 2023.

\bibitem[Tsai et~al.(2023{\natexlab{b}})Tsai, Liu, and Ren]{tsai2023rtlfixer}
YunDa Tsai, Mingjie Liu, and Haoxing Ren.
\newblock Rtlfixer: Automatically fixing rtl syntax errors with large language models.
\newblock \emph{arXiv preprint arXiv:2311.16543}, 2023{\natexlab{b}}.

\bibitem[Addison~Howard(2018)]{petfinder-adoption-prediction}
Mongrel~Jedi Addison~Howard, MichaelApers.
\newblock Petfinder.my adoption prediction, 2018.
\newblock URL \url{https://kaggle.com/competitions/petfinder-adoption-prediction}.

\bibitem[Da~Tsai and De~Lin(2022)]{da2022fast}
Yun Da~Tsai and Shou De~Lin.
\newblock Fast online inference for nonlinear contextual bandit based on generative adversarial network.
\newblock \emph{arXiv preprint arXiv:2202.08867}, 2022.

\bibitem[Peng et~al.(2023{\natexlab{b}})Peng, Wang, Dong, Hao, Huang, Ma, and Wei]{peng2023kosmos}
Zhiliang Peng, Wenhui Wang, Li~Dong, Yaru Hao, Shaohan Huang, Shuming Ma, and Furu Wei.
\newblock Kosmos-2: Grounding multimodal large language models to the world.
\newblock \emph{arXiv preprint arXiv:2306.14824}, 2023{\natexlab{b}}.

\bibitem[Alayrac et~al.(2022)Alayrac, Donahue, Luc, Miech, Barr, Hasson, Lenc, Mensch, Millican, Reynolds, et~al.]{alayrac2022flamingo}
Jean-Baptiste Alayrac, Jeff Donahue, Pauline Luc, Antoine Miech, Iain Barr, Yana Hasson, Karel Lenc, Arthur Mensch, Katherine Millican, Malcolm Reynolds, et~al.
\newblock Flamingo: a visual language model for few-shot learning.
\newblock \emph{Advances in Neural Information Processing Systems}, 35:\penalty0 23716--23736, 2022.

\bibitem[Tsai et~al.(2021)Tsai, Chen, and Lin]{tsai2021toward}
Yun-Da Tsai, ChengKuan Chen, and Shou-De Lin.
\newblock Toward an effective black-box adversarial attack on functional javascript malware against commercial anti-virus.
\newblock In \emph{Proceedings of the 30th ACM International Conference on Information \& Knowledge Management}, pages 4165--4172, 2021.

\end{thebibliography}
\clearpage

\appendix
\onecolumn

\section{Appendix}
\label{sec:appendix}

In Section~\ref{sec:analysis}, we analyze the data after summary and reasoning, and we only highlight the most important parts. In this section, we will present the detailed raw input for a more comprehensive analysis.

\subsection{Noise in Text Data}
\label{sec:appendix:text}

\begin{figure*}[ht]

\begin{tcolorbox}[width=1.0\linewidth, halign=left, colframe=black, colback=white, boxsep=0.01mm, arc=1.5mm, left=2mm, right=2mm, boxrule=0.5pt]\footnotesize
\textbf{Text transformation - text}\\ 

The type of pet is Dog. The name of pet is Jack Jack. The age of pet when listed (in months) is 36. The gender of pet is Male. The color 1 of pet is Brown. The color 2 of pet is Unknown. The color 3 of pet is Unknown. The size of pet at maturity is Small. The fur length of pet is Short. The vaccination status of pet is Yes. The deworming status of pet is Yes. The sterilisation status of pet is No. The health condition of pet is Minor-Injury. The number of pets represented in profile is 1. The pet adoption fee is Free. The state location of pet in Malaysia is Selangor. The total uploaded videos for this pet is 0. The total uploaded photos for this pet is 3. The primary breed of pet is Chihuahua. The secondary breed of pet (if pet is of mixed breed) is Unknown.

The image presents a small, fawn-colored dog with large ears that stand upright, creating an alert and inquisitive expression. The dog is looking up towards the camera with its head slightly tilted, which may indicate curiosity or attentiveness. It is wearing a light blue collar with a bell, signifying that it is a pet likely kept indoors. The dog is positioned on a plain tiled floor, with the low angle perspective of the shot highlighting its petite stature and endearingly disproportionate ear-to-head size ratio.

\textcolor{red}{The profile write-up for this pet is : Jack experienced it Jack dog even is He recovering but special care skin which hair to Vet this was to Hopefully lover who take good dog a}
\end{tcolorbox}

\begin{tcolorbox}[width=1.0\linewidth, halign=left, colframe=black, colback=white, boxsep=0.01mm, arc=1.5mm, left=2mm, right=2mm, boxrule=0.5pt]\footnotesize

\textbf{After Summarization and Reasoning - text}\\ 
This pet is a charming 36-month-old male Chihuahua named Jack Jack, who has a short brown fur coat. Despite a minor injury, he is in good health and has been both vaccinated and dewormed, although he has not been sterilized. Jack Jack is a small-sized dog who currently resides in Selangor, Malaysia, and is available for adoption free of charge. Jack Jack's profile includes three photos, one of which captures him as a small, fawn-colored dog with large, upright ears that give him an alert and inquisitive look. He is seen wearing a light blue collar with a bell, suggesting he is an indoor pet. The image shows him on a plain tiled floor, with his head slightly tilted up towards the camera, highlighting his petite stature and the endearing size ratio of his ears to his head.
\textcolor{blue}{Although the profile write-up for Jack Jack is somewhat fragmented, it indicates that he has experienced some challenges but is on the road to recovery. He requires special care for his skin and fur, and hopefully, a dog lover will come forward to provide him with the good care he deserves.} For more information or photos, interested adopters can reach out via Whatsapp. Several factors influence the adoption rate for Jack Jack, and while he has some positive attributes, there are also some challenges that may slow down his adoption.

1. **Breed and Size**: Jack Jack is a Chihuahua, a breed that is generally popular due to their small size and suitability for apartment living. This is a positive factor that could help in his adoption.

2. **Age**: At 36 months (3 years old), Jack Jack is no longer a puppy. Puppies tend to get adopted faster than adult dogs, which may slow down his adoption rate.

\textcolor{cyan}{3. **Health Condition**: Jack Jack has a minor injury and is recovering. Potential adopters might be hesitant to take on a pet that requires special care, even if the injury is minor.}

4. **Vaccination and Deworming**: The fact that Jack Jack is vaccinated and dewormed is a positive aspect and can reassure potential adopters about his health.

5. **Sterilisation Status**: Jack Jack is not sterilized, which could be a concern for some adopters. Many adopters prefer pets that are already neutered to avoid the cost and responsibility of the procedure.

6. **Profile Write-Up**: The profile write-up for Jack Jack is somewhat unclear and lacks detailed information that could appeal to potential adopters. A well-written profile can significantly impact adoption rates.

7. **Photos**: There are only 3 photos uploaded. More photos and possibly videos could help showcase Jack Jack's personality and increase interest.

8. **Location**: Jack Jack is located in Selangor, Malaysia, which has a significant population and potentially a larger pool of adopters. However, the adoption rate can still vary based on local preferences and the number of available pets.
\end{tcolorbox}
\caption{LLM summarization utilizes its capabilities to infer the original meaning of sentences obscured by noise. LLM Reasoning further transforms implicit relations into explicit text descriptions. (red: noisy text, blue: results inferred through summarization, light blue: potential meaning behind the information uncovered through reasoning.) This example demonstrates the effectiveness of our text-centric methods.}
\label{fig:appendix-analysis-text}
\end{figure*}

\begin{figure*}
\subsection{Noise in Tabular Data}
\label{sec:appendix:image}
\end{figure*}

\begin{figure*}[ht]

\begin{tcolorbox}[width=1.0\linewidth, halign=left, colframe=black, colback=white, boxsep=0.01mm, arc=1.5mm, left=2mm, right=2mm, boxrule=0.5pt]\footnotesize

\textbf{Text transformation - table}\\ 
\textcolor{gray}{The type of pet is Dog. The name of pet is Filo. \textcolor{red}{The age of pet when listed (in months) is 78.} The gender of pet is Male. The color 1 of pet is Brown. The color 2 of pet is Unknown. The color 3 of pet is Unknown. The size of pet at maturity is Large. The fur length of pet is Short. The vaccination status of pet is Yes. The deworming status of pet is Yes. The sterilisation status of pet is Unsure. The health condition of pet is Healthy.\textcolor{red}{ The number of pets represented in profile is 1.} The pet adoption fee is Free. The state location of pet in Malaysia is Kuala Lumpur. The total uploaded videos for this pet is 0.\textcolor{red}{The total uploaded photos for this pet is 5.} The primary breed of pet is Mixed Breed. \textcolor{red}{The secondary breed of pet (if pet is of mixed breed) is Unknown}}

\textcolor{violet}{The image shows a dog with a light brown, short coat lying comfortably on a blue plastic crate.} The dog has a gentle expression, with soft brown eyes, and appears to be calmly resting inside what looks like a tiled room, likely a part of its home. The dog's collar with an identification tag suggests it is well-loved and cared for. The environment looks clean and provides a sense of a safe and secure space for the pet to relax.

The profile write-up for this pet is : Meet Filo, the cute and handsome 6 year old mixed breed boy. Filo is a lucky stray saved by a kind owner earlier this year. \textcolor{violet}{This kind owner brought Filo to the vet to treat some external wounds.} However, the kind owner has 2 dogs and don't have room for Filo. Filo continues to stay in the cage with the vet from that day till now while his wounds are treated. When the vet is available, Filo gets to go out for a short walk. If the clinic is packed, Filo stays in the cage for the whole day. Its not anybody's fault. Filo is still grateful for this kind owner and the kind vet for their hospitality. But Filo's future is still very uncertain and he is hoping that he's lucky to be granted a permanent loving home. I met Filo when I brought my poodle to visit the clinic. I learn of Filo's story from this kind vet. If you can give Filo a loving home or if you need more information, please call me. 

\end{tcolorbox}

\begin{tcolorbox}[width=1.0\linewidth, halign=left, colframe=black, colback=white, boxsep=0.01mm, arc=1.5mm, left=2mm, right=2mm, boxrule=0.5pt]\footnotesize

\textbf{After Summarization and Reasoning - table}\\ 
\textcolor{blue}{This pet is a gentle and handsome 6-year-old mixed breed dog named Filo. He has a light brown, short coat and soft brown eyes,} giving him a calm and serene expression. Filo's story is one of resilience and hope. He was a stray who was fortunate to be rescued by a kind owner earlier this year. This compassionate person took Filo to the vet to treat some external wounds, but due to having two other dogs, they couldn't keep Filo at home. Since then, Filo has been staying at the vet's clinic, where he continues to receive care and treatment for his wounds.

Filo's life at the clinic is a mix of confinement and occasional freedom; he stays in a cage most of the day but gets to go out for short walks when the vet is available. Despite these limitations, Filo remains grateful for the kindness shown to him by both his rescuer and the vet. However, his future remains uncertain, and he longs for a permanent, loving home where he can feel safe and secure.

Filo's profile features five photos, one of which shows him lying comfortably on a blue plastic crate inside a tiled room, likely part of his temporary home at the clinic. The clean and secure environment, along with his collar and identification tag, suggests that he is well-loved and cared for. If you can provide Filo with the forever home he deserves or need more information, please call the number provided. Filo's adoption rate is likely to be more than 3 months for several reasons:

1. **Age**: At 78 months (6.5 years old), Filo is significantly older than the other pets listed. Younger pets, especially puppies and kittens, generally have higher adoption rates because they are perceived as more adaptable and have a longer potential lifespan.

\textcolor{blue}{2. **Health and Background**: While Filo has been treated for his external wounds, his background as a stray and his extended stay in a cage might raise concerns for potential adopters about his health and behavior. The write-up mentions his gratefulness and calm demeanor, but it does not provide detailed information about his current health status or behavior, which could be crucial for potential adopters.}

3. **Emotional Connection**: The profile write-up is heartfelt and provides a touching backstory, but it lacks the emotional appeal seen in the profiles of younger pets. The language used to describe Filo's situation is more factual and less engaging compared to descriptions of other pets, which emphasize their cuteness and playful nature.

4. **Visual Appeal**: Although there are 5 photos, the description of the image shows Filo in a resting position, which may not be as engaging as images of playful or interactive behavior. Potential adopters often respond more positively to images that show the pet's personality and energy.

5. **Competition**: Filo is competing with younger, more visually appealing pets that are often adopted faster. His profile needs to stand out more to attract potential adopters who are specifically looking for an older, more mature dog.

6. **Location and Accessibility**: The profile does not specify the exact location beyond being at a vet clinic, which might make it less accessible for potential adopters who prefer to know more about where the pet is currently staying.
\end{tcolorbox}

\caption{An example of LLM summarization recovering noisy tabular data through other modality information.(gray: dropped columns, red: retained column, purple: data referenced by the summarization, blue: data generated after compensation missing tabular columns information from image.) In this example, even after dropping table information such as the pet's health condition, fur length, and color, these details can still be recovered from images and text descriptions.}
\label{fig:appendix-analysis-table}
\end{figure*}

\end{document}